\documentclass{article}
\usepackage{ijcai20}

\usepackage{times}

\usepackage{soul}
\usepackage{url}
\usepackage[hidelinks]{hyperref}
\usepackage[utf8]{inputenc}
\usepackage[small]{caption}
\usepackage{graphicx}
\usepackage{placeins}
\usepackage{amsmath}
\usepackage{adjustbox}
\usepackage{footmisc}
\usepackage{booktabs}
\usepackage{multirow}
\usepackage{stackengine}
\usepackage{xcolor}

\newcommand{\resnext}{ResNeXt-IG-3.5B}

\title{All-in-One Image-Grounded Conversational Agents}

\author{
Da Ju\footnote{Contact Author}\and
Kurt Shuster \and
Y-Lan Boureau  \And
Jason Weston \\
\affiliations
Facebook AI Research \\
\emails
\{daju, kshuster, ylan, jase\}@fb.com
}

\begin{document}

\maketitle

\begin{abstract}
As single-task accuracy on individual language and image tasks has improved substantially
in the last few years, 
the long-term goal of a generally skilled agent that can both see and talk becomes more feasible to explore.
In this work, we focus on leveraging individual language and image tasks, along with
resources that incorporate {\em both} vision and language towards that objective. We design an architecture that
combines state-of-the-art Transformer and ResNeXt modules fed into a novel attentive multimodal module 
to produce a combined model trained on many tasks. 
We provide a thorough analysis of the components of
the model, and transfer performance when training on one, some, or all of the tasks.
Our final models provide a single system that obtains good results
on all vision and language tasks considered, and  improves the state of the art in image-grounded 
conversational applications.
\end{abstract}

\section{Introduction}

A picture may be worth a thousand words, but combining pictures and words is even better.
There are many ways to marry vision and language: an image can be a great conversation starter, or discussion point;
text accompanying an image can be a mere descriptive caption,
or some witty commentary. 
Humans can seamlessly blend these skills and use them for interaction, depending on the given setting and need.

In order to probe this  range of skills, a large
set of image-and-text tasks have been devised by researchers, covering image captioning \cite{young2014image,chen2015microsoft,shuster2019engaging},
visual question answering \cite{balanced_vqa_v2,das2017visual}, and dialogue based on an image \cite{igc,shuster2018engaging}. 
Recent years have seen tremendous progress in both vision \cite{he2016deep,Detectron2018,uru} and language \cite{vaswani2017attention,radford2019language,devlin-etal-2019-bert} applications, and in all
these individual tasks \cite{balanced_vqa_v2,shuster2018engaging,shuster2019engaging} as well, so the time is ripe for exploring the possibility of a multi-tasking agent that would do well on all of them.

In this work, we design an architecture that leverages existing state-of-the-art vision and language modules, and combines them with a novel attentive multimodal combiner module. 
The module can learn when and how to attend between the two modalities, depending on the particular inputs, and improves over a standard attention mechanism.
Our work also provides a detailed analysis of what works and what does not.
We perform multiple ablation
experiments to compare what types of architectures, training objectives, and optimization strategies
work best for what tasks, or for achieving the most balanced performance across the board.
We thus obtain models that improve the state of the art over several individual image-grounded conversational tasks, and a
single system that is capable of doing well on all the image-grounded language tasks we consider.

\begin{table*}   
\begin{center}
\begin{small}
\begin{tabular}{c|l|cc|cc|cc|c}
\hline
\multirow{2}{*}{Modalities} & 
\multirow{2}{*}{Task} & \multicolumn{2}{c|}{Train} & \multicolumn{2}{c|}{Valid} & \multicolumn{2}{c|}{Test} &  \multirow{2}{*}{\# Cands}\\
 && \# Images & \# Utterances & \# Images & \# Utterances & \# Images & \# Utterances & \\
\toprule
\multirow{2}{*}{Language} & Wiki. + Tor. Books & - & 150m & - & - & - & - & -\\
& pushshift.io Reddit & - & 174m & - & - & - & - & -\\
\hline
\multirow{3}{*}{Vision} & ImageNet& 1.28m & - & - & - & - & - & -\\
& Instagram & 3.5b & - & - & - & - & - & -\\
& Visual Genome & 108,077 & - & - & - & - & - & -\\
\hline
\multirow{3}{*}{Vision} & COCO & 82,783 & 414,113 & 5,000 & 25,000 & 5,000 & 25,000 & 5,000 \\
& Flickr30k & 29,000 & 145,000 & 1014 & 5,070 & 1,000 & 5,000 & 1,000 \\
& Personality-Captions & 186,858 & 186,858 & 5,000 & 5,000 & 10,000 & 50,000 & 500\\
\multirow{1}{*}{ +} & Image-Chat & 186,782 & 355,862 & 5,000 & 15,000 & 9,997 & 29,991 & 100\\
\multirow{3}{*}{Language} & Image-Chat QA & 19,702 & 19,702 & 1,129 & 1,129 & 2,224 & 2,224 & 100\\
& IGC &  - & - & 1,613 & 4,839 & 2,591 & 7,773 & 100\\
& VQA & 82,783 & 443,757 & 40,504 & 214,354 & 81,834 & 447,793 & 3,129\\
\bottomrule
\end{tabular}
\end{small}
\end{center}
\caption{Dataset statistics for all relevant datasets. During evaluation, gold responses are scored against other candidates (\#Cands).
\label{table:datasets}
}
\end{table*}

\section{Tasks}

We first detail separate language and vision tasks that are considered from prior work,
and then describe the combined vision and language tasks we consider for training an entire architecture for building an image-grounded conversational agent. 
A summary of these tasks is also provided in  Table~\ref{table:datasets}.

\subsection{Language-based}

Large-scale text corpora are commonly used to pre-train text encoders;
we use these methods that have been developed in  prior work.
In particular we first consider BERT-based representations \cite{devlin-etal-2019-bert} from \cite{humeau2019real},  which use 150 million (context, response) pairs extracted from Wikipedia and Toronto Books. To make use of data potentially
more related to dialogue and of a more colloquial nature, we also use pre-training based on  pushshift.io
Reddit \cite{mazare2018training,humeau2019real}, consisting of 174 million (context, response) pairs.

\subsection{Vision-based}

Similarly, large-scale image datasets are commonly used to pre-train image encoders,
in particular ImageNet \cite{deng2009imagenet} (1.28 million images), Instagram images \cite{uru} (3.5 billion images),
and the Visual Genome (108k Images with 2.8 Million attributes) \cite{krishnavisualgenome}.

\subsection{Vision + Language}\label{sec:vltasks}

In the combined tasks we consider, images and language are possible inputs, and the output is a text response from the agent. The goal is that the tasks, when multi-tasked,
can teach
an agent how to respond appropriately in different situations using different skills.

\paragraph{COCO Captions}

The COCO Captions dataset \cite{chen2015microsoft} requires that a model, given an image, predicts a caption that factually summarizes the scene, for example ``a large bus sitting next to a very tall building''.
In the dataset used for the 2015 challenge, there are about 83k training images and 414k captions, as images are captioned multiple times, and a large validation set of about 40k images.
Some works have merged
some or all images from that validation set into the training set (we indicate this with an asterisk in Table~\ref{table:previous-work}). 
In this work, we only train on the 83k images of the original train set, to avoid training on images that also appear in the VQA validation set,
and use the validation and test sets of 5k images each from \cite{2017karpathycocosplit}.

\paragraph{Flickr30k}
Flickr30k \cite{young2014image} is also a captioning dataset with factual summaries, although it is smaller with 29k training images and 145k captions.

\paragraph{Personality Captions (PC)}
In contrast to the previous two datasets, Personality Captions \cite{shuster2019engaging} attempts to model human style when people speak about images. While the training set also consists of (image, response) pairs, each one also has a given style label out of 215 possible styles, such 
as ``Sympathetic'', ``Optimistic'' or  ``Dramatic''. The captions authored by humans then tend to be less factual in tone, and rather than simply stating what is in the image they are  more conversational, e.g. ``This sandwich looks so delicious! My goodness!''.
It consists of about 187k training images, with one caption each.

\paragraph{Image Chat (IC)}
Image Chat \cite{shuster2018engaging} is an extension of the Personality Captions dataset to full dialogue. It also uses the same 215 style traits and images as input, but human-human conversations have been collected based on the images and traits instead, with each speaker pair in a given chat being assigned a possibly different random trait.  The training set consists of 
the same 187k images with 356k total conversation turns.

\paragraph{Image Chat QA (ICQA)}
Image Chat QA is the extraction of all the question-answer pairs that appear in the Image Chat dataset, to evaluate performance in answering such conversational image-grounded questions. The questions have been extracted heuristically, by assuming a question contains a {\em ?} or starts with {\em who}, {\em what}, {\em when}, {\em where}, {\em why} or {\em how}. This extracts about 20k such training questions.

\paragraph{Image-Grounded Conversations (IGCQ and IGCQA)}
Image-Grounded Conversations (IGC) \cite{igc} is also a conversational dataset between pairs of humans given an image. It does not contain a training set, but only validation and test portions. 
The conversations are three turns each, in the format of (context, question, response) tuples. We refer to the task of forming a question given the context as IGCQ, and the task of responding to the question as IGCQA.

\paragraph{VQA}

Visual QA \cite{balanced_vqa_v2} is a task involving open-ended questions about images which require an understanding of vision, language, and commonsense knowledge to answer, such as ``where is the child sitting?"' or ``who is wearing the glasses?''. It contains 83k training images and 444k QA pairs.
Note this line of work has also been extended to multiple questions in sequence \cite{das2017visual} but we do not consider that task here.

\section{Related Work}\label{sec:related}

Separately in the NLP field, and in the vision field, large advancements have been recently
made in terms of the quality of learnt representations.

In NLP, word embedding representations \cite{bengio2003neural,collobert2008unified,mikolov2013distributed,fasttext}
have given way to multi-sentence, multi-layer, self-attentive 
representations through Transformers, with pre-training on large corpora such as Wikipedia 
and Toronto books
\cite{vaswani2017attention,radford2018improving,devlin-etal-2019-bert,bakhtin2019real,radford2019language}. In dialogue, it has been shown that pre-training on utterances from large-scale
conversational corpora
such as from pushshift.io Reddit improves over large pre-training over resources like Wikipedia because they are more related to the task \cite{mazare2018training,humeau2019real,shuster2019dialogue}. When training on downstream tasks, multi-tasking language tasks is also starting to become a more explored area \cite{collobert2008unified,mccann2018natural,raffel2019exploring}.

In vision, conventional convolutional neural networks \cite{lecun1990handwritten,krizhevsky2012imagenet} have
been upgraded and improved by deeper ResNet architectures that incorporate skip connections \cite{he2016deep}, trained through ImageNet \cite{deng2009imagenet}. 
On tasks such as VQA which explicitly ask questions about object properties, Faster R-CNN features \cite{Detectron2018}, which incorporate object detection algorithms, have been shown to perform well. On tasks with large coverage of everyday images and commonsense knowledge about them, Instagram training has been shown to perform well \cite{uru,shuster2018engaging,shuster2019engaging}.

Given this improved performance across different modalities, a natural next step is methods that combine these approaches for multimodal tasks involving language and vision. 
Several recent approaches have been built with this goal, including Vilbert \cite{lu2019vilbert}, VisualBERT \cite{li2019visualbert}, LXMERT \cite{tan2019lxmert}, 
Unicoder-vl \cite{li2019unicoder}, Vl-bert \cite{su2019vl} and UNITER \cite{chen2019uniter}.
A common theme is to borrow some of the pre-training ideas from BERT, but apply them to
pre-training both language and vision, and then fine-tune these models on downstream
tasks. Another recent work multi-tasks 12 vision and language tasks at once \cite{lu201912}.
Somewhat differing from our work, the end tasks considered are not to aimed to build a unified conversational agent where the output is dialogue, but include any task including language and vision of some form,
for example caption-based image-retrieval, referring expressions and region to phrase grounding, most of which we do not consider here.  Recently, \cite{shuster2019dialogue} proposed to multi-task to build a conversational agent, but using mostly language-only tasks (10 tasks), although it does include two of the image tasks we consider here.


\section{Methods}
\begin{figure}[htp]
    \centering
    \includegraphics[width=8cm]{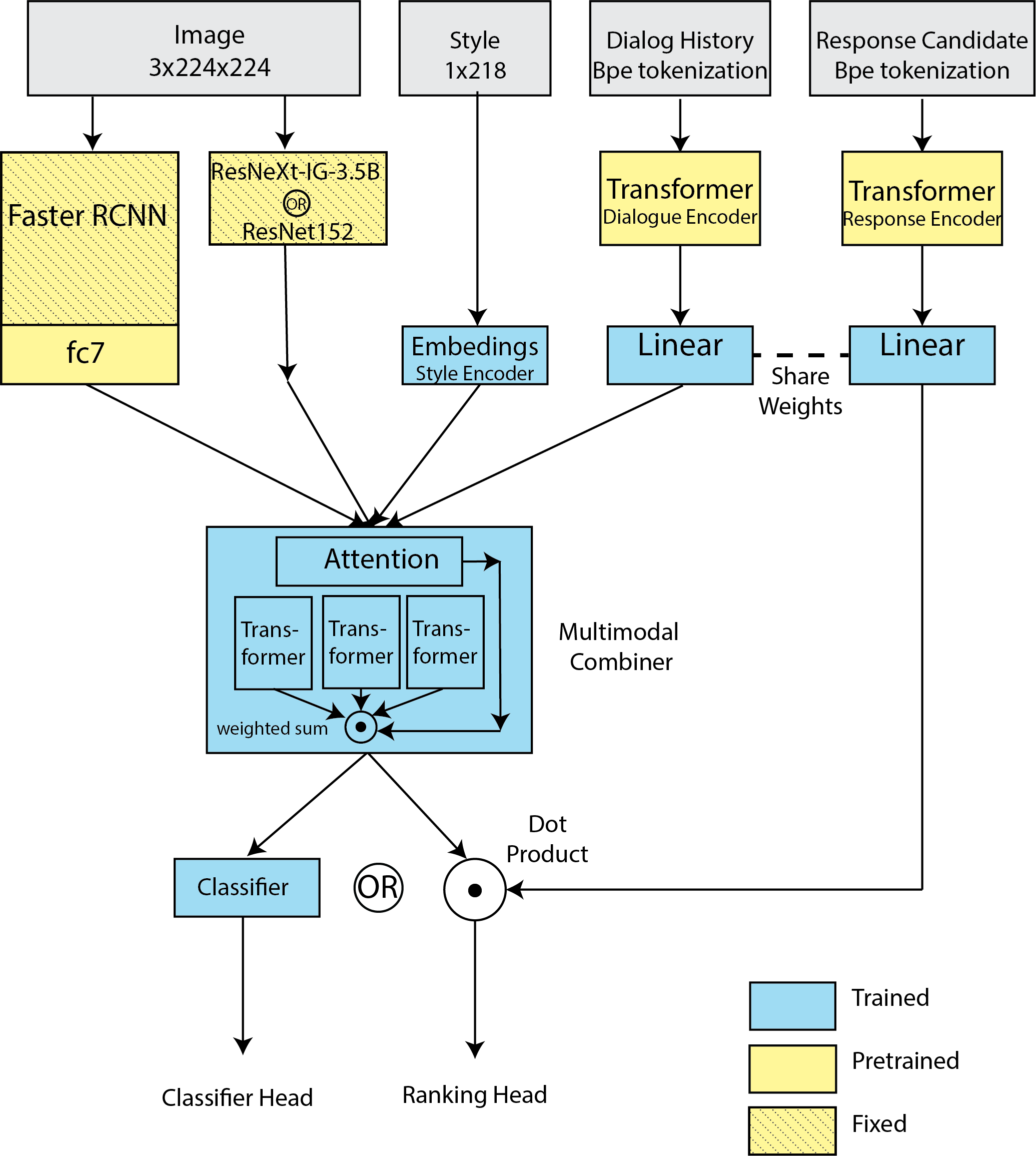}
    \caption{Overview of our model, in its  TransResNet-3AMMC (Attentive Multimodal combiner) variant. The non-attentive variant (TransResNet-MMC) has a single Transformer combiner, instead of the three split Transformers followed by a weighted sum shown here.}
    \label{fig:galaxy}
\end{figure}

Our model is a retrieval architecture that outputs a candidate response from the training set. Like most multimodal architectures, it comprises a text encoder, an image encoder, and a way to combine the two. However, unlike recent models that use various cross attention mechanisms to get the joint representation of the final context, our model simply uses a so-called multimodal combiner. An extra style encoder is also added to represent the different style traits inside the Personality Captions and Image Chat tasks, and to differentiate them from the other tasks. The model finally scores possible output candidates, using either a ranking or a classification head, depending on the task.
An overview of the model architecture is given in Figure ~\ref{fig:galaxy}.

\subsection{Text Encoders} \label{sec:text-enc}
We use two text encoders, one for the input context and one for the output candidates. The output candidate encoder encodes the candidate utterances that will be scored, and the context encoder encodes the text input. Depending on the task, the context can be the previous dialogue history or a question posed about the image, or a combination of the two. 
Both text encoders are pre-trained Transformers. 
The final output of the candidate encoder is a single vector per candidate, obtained by taking the mean of the per-token encodings. For the context encoder we retain the per-token output encodings, thus comprising the same length as the input sequence, for input into the multimodal combiner. 
During multimodal training we fine-tune both text encoders.

\subsection{Image Encoder} \label{sec:image-enc}
We consider two possible image encoders in our model.
The first is a ResNeXt-based model trained on 3.5 billion Instagram images \cite{uru}, which we refer to as \resnext. This encodes the image into a 2048-dimensional vector.
The weights are fixed during the subsequent training process.
The second is an improved Faster R-CNN model \cite{Detectron2018} trained on the Visual
Genome dataset \cite{krishnavisualgenome}. We fix the network up to the fc6 layer and fine-tune the fc7 weights as in \cite{pythia01}. 
We extract 100 2048-dimensional vectors (100-channel Faster R-CNN features).
In addition to trying these models independently, we also investigate using both of their features concatenated together as input to the multimodal combiner.

\subsection{Multimodal Combiner} \label{sec:mmc}
The Multimodal Combiner (MMC) is used to combine the encodings from different components of the model. This includes the 100-channel Faster R-CNN features, the \resnext~features, the sequence-based context features (which depend on the text length), and the encoding from the style encoder. 
Prior to combination, the individual encodings are normalized with their own layer-norm layer; each is then fed into the MMC with a positional embedding to indicate which feature type it is.
The Multimodal Combiner is simply a 
Transformer
\cite{attentionisallyourneed} encoder without its embedding layer; thus, self-attention is applied to all features that then go through linear layers. A mean operation is performed in the end to get a single vectorial representation of the whole multimodal context. This joint representation is then either used for a dot product with the candidate encodings (for ranking) or sent to an additional linear layer (for classification), as detailed in Sec.~\ref{sec:heads}.

\subsection{Attentive Multimodal Combiner} \label{sec:ammc}
We further propose an Attentive Multimodal Combiner (AMMC), shown in Fig.~\ref{fig:galaxy}, where multiple Transformers are used and then combined through an attention mechanism, potentially allowing them to focus on different skills. We use the style encoding as the query, forward it to a linear layer of the same output dimension as the number of Transformers in the multimodal combiner (i.e., 2 to 4, denoted as 2AMMC, 3AMMC and 4AMMC),
followed by a softmax. 
We hence use those outputs to perform a weighted sum of the outputs from all the multimodal Transformers. This attention mechanism thus learns which Transformers to rely on more for which inputs, allowing the model to switch between skills.

\subsection{Output Heads and Loss} \label{sec:heads}

For tasks like VQA where there is one factually correct answer and many wrong answers, it has been shown that strong performance can be achieved using a
{\em classification head}, considering all the possible most frequent answers as classes, and using a binary cross entropy loss\footnote{As used in Pythia \cite{singh2019TowardsVM,singh2018pythia} (\url{https://github.com/facebookresearch/pythia}).}. 
We thus also consider this approach for VQA.
For open-ended problems in which there may be multiple right answers, e.g. those in 
Image Chat, we consider an alternative approach of using a ranking head.
In this approach, 
the gold label is contrasted with a subsample of negative candidates during training (chosen as the labels of other examples in the batch) but still using the  binary cross entropy loss, 
which  scales well to huge candidate sets. We compare these two methods in this work, and also considering training both at the same time.
We use batch sizes of 256 / 512 and adam for optimization.
For multi-tasking we experimented with various kinds of dataset weighting schemes, but in the end we went for simplicity and report results of sampling from tasks equally, so that the same number of updates are done per task in an epoch, which was difficult to improve upon.

\begin{table*}[ht!]
\begin{center}
\begin{small}
\begin{tabular}{c|c|ccccc}
\hline
\multicolumn{1}{c}{Image Encoder} &    & COCO           & Flickr30k     & Image Chat      & VQA         & Avg            \\ 
\toprule
\multirow{2}{*}{\resnext}   & ST & 50.7          & 75.3          & \textbf{56.4} & 61.9       & 61.1          \\ 
                     & MT & 48.0          & 77.0            & 56.2          & 62.0       & 60.8          \\ 
\midrule
\multirow{2}{*}{Faster R-CNN}   & ST & 49.3           & 68.2          & 54.2          & 66.3       & 59.5          \\  
                     & MT & 52.1          & 72.4          & 53.2          & 66.3        & 61.0          \\ 
\midrule
\multirow{2}{*}{\resnext + Faster R-CNN} & ST & \textbf{57.3} & 79.7          & \textbf{56.4}          & \textbf{67.0} & \textbf{65.1} \\
                     & MT & 51.2          & \textbf{81.7} & 55.2           & 66.4       & \textbf{63.7} \\ 
                     \bottomrule
\end{tabular}
\end{small}
\end{center}
\caption{Comparison between image representations as part of our TransResNet-MMC architecture with either single-task (ST) or multi-task (MT) training, evaluating on COCO, Flickr30k, Image Chat and VQA, and reporting average (Avg.) performance across the tasks.
\label{table:image-rep}
}
\end{table*}

\begin{table}   
\begin{center}
\begin{small}
\begin{tabular}{l|cccc}
\hline
Text Encoder & COCO & Flickr30k & Image Chat & Avg.\\
\toprule
{\em from scratch} & 40.7 & 65.5 & 37.6 & 48.0 \\
fastText init  & 44.9 & 69.0 & 45.6 & 53.2 \\
BERT           & \textbf{50.1} & \textbf{72.0} & 52.1  & 58.1\\
Reddit-79M &44.3 & 68.4 & 50.3  & 54.3 \\
Reddit-128M 
   & 48.8 & 71.8 & \textbf{55.2} & \textbf{58.6}\\
\bottomrule
\end{tabular}
\end{small}
\end{center}
\caption{Comparison between text encoding Transformer pre-training methods  when used as part of TransResNet-MMC, reporting accuracy on the respective test sets of three tasks, as well as the average (Avg.).
\label{table:text-rep}
}
\end{table}

\begin{table}
\begin{small}
\begin{center}
\begin{tabular}{c|cccccc}
\hline
 Early Stop       & COCO           & Fl30k         & PC             & IC            & ICQA    & VQA            \\ \toprule
COCO    & \textbf{54.0} & \textbf{83.4} & 55.0          & 50.5         & 43.9          & 66.1          \\ 
Fl30k  & 51.4          & 83.0          & 55.9           & 53.1         & 47.2          & 60.3          \\ 
IC      & 52.4          & 81.3           & \textbf{58.8} & \textbf{55.9} & \textbf{51.4} & 66.5           \\ 
VQA     & 53.4          & 81.9           & 58.0          & 54.0         & 30.6          & \textbf{66.6} \\ 
Avg. & 51.2          & 81.7           & 58.0          & 55.2          & 49.9           & 66.4          \\ 
\bottomrule
\end{tabular}
\end{center}
\end{small}
\caption{Training TransResNet-MMC on all tasks but only performing early stopping on one specific dataset compared to stopping on the average accuracy across all datasets (``Avg.'').
\label{table:mt-st-earlystop}
}
\end{table}

\begin{table}
\begin{center}
\begin{small}
\begin{tabular}{c|cccccc}
\hline
Fine Tune & COCO          & Fl30k      & PC         & IC    & ICQA        & VQA            \\ \toprule
COCO      & \textbf{59.6} & 76.5        & 34.0      & 31.8 & 30.0      & 58.2          \\
Flickr30k    & 50.7         & \textbf{84.0} & 54.2       & 52.1  & 47.1       & 60.8           \\
IC        & 52.4 & 81.3 & \textbf{58.8} & \textbf{55.9} & \textbf{51.4}	& \textbf{66.5}	\\
VQA       & 36.6  & 65.6 & 47.1 & 38.6 & 30.7 & 66.2 \\
All   & 51.2         & 81.7        & 58.0      & 55.2  & 49.9       & 66.4 \\
\bottomrule
\end{tabular}
\end{small}
\end{center}
\caption{Training TransResNet-MMC on all tasks and then fine-tuning on each of the tasks, compared to the original best performing multi-task model (called ``All'').
\label{table:mt-plus-ft}
}
\end{table}

\begin{table}
\begin{center}
\begin{small}
\begin{tabular}{l|ccccc}
\hline
VQA Model training             &  class. head & ranking head         \\ \toprule
Classification head    & 67.0  & n/a \\
\midrule
Ranking head     &  n/a  & 54.0 \\
\midrule
Multi-head training    & 66.1  & 63.5 \\
\bottomrule
\end{tabular}
\end{small}
\end{center}
\caption{
Training VQA with either a classification head, a ranking head, or multi-tasking both. Multi-tasking both helps the ranking head improve.
\label{table:heads}
}
\end{table}

\begin{table}
\begin{center}
\begin{small}
\begin{tabular}{cr|ccc}
\hline
 && COCO & Fl30k & IC\\
\toprule
\multirow{2}{*}{ResNeXt-IG-3.5B} &w/o MMC & 48.8 & 71.8 & 55.2 \\
& with MMC & 50.7 & 75.3 & 56.6  \\
\midrule
{ResNeXt-IG-3.5B} &w/o MMC & 53.6 & 75.6 & 46.9 \\
+ Faster R-CNN&with MMC & 57.3 & 79.7 & 56.4  \\
\bottomrule
\end{tabular}
\end{small}
\end{center}
\caption{Comparison of with and without (w/o) the  multimodal combiner (MMC) as part of our TransResNet architecture, for COCO, Flickr30k (Fl30k) and Image Chat (IC),
using either \resnext~(ResNeXt-IG) features alone or in combination with Faster R-CNN features.
The MMC provides gains in all cases.
\label{table:mix-rep}
}
\end{table}

\section{Experiments}

We now describe our experiments, in which we 
perform analysis and ablations of the different kinds of modules and inputs we use for training, and final results on our full architecture.
For all models we choose hyperparameters on the validation set(s), and report on the test set; for VQA the numbers are reported on the test-dev set.
All experiments were conducted in ParlAI \cite{miller2017parlai}, and we plan to make the code publicly available.

\paragraph{Text Encoding}

We consider different possible text encodings with different pre-training schemes:
starting from random weights before training on our language + vision tasks;
starting from initialized word embeddings from fastText \cite{fasttext} only; starting
from BERT weights \cite{devlin-etal-2019-bert}; 
and starting from two versions of pushshift.io Reddit training, i.e., Transformers with 79M parameters from \cite{mazare2018training} and 128M parameters from \cite{humeau2019real}.  
After initialization we then fine-tune the entire TransResNet-MMC, using \resnext~image features, on three tasks separately: COCO, Flickr30k and Image Chat.
The results are given in Table \ref{table:text-rep}.

We observe large improvements in accuracy with more text pre-training, for example on COCO going from  40.7\% with no pre-training to 50.1\% with BERT.
BERT outperforms pushshift.io Reddit-128M  slightly on COCO and Flickr30k, whereas pushshift.io Reddit-128M outperforms BERT on Image Chat. We hypothesize this is because the language is more formal on COCO and Flickr, matching BERT that is trained with Wikipedia, whereas Image Chat is more colloquial, matching pushshift.io Reddit. However, the results are close and on average pushshift.io Reddit-128M does slightly better.
We thus use the latter in all subsequent experiments\footnote{Another choice would have been to combine them, but we did not do that here.}.

\paragraph{Image Encoding}

We next consider different possible image encodings via different architectures and pre-training schemes:
\resnext~\cite{uru}, 
Faster R-CNN features \cite{Detectron2018}, 
and finally a combination of \resnext~ and Faster R-CNN features. 
After initialization we then fine-tune the entire TransResNet-MMC on four tasks:
COCO, Flickr30k, Image Chat and VQA. We evaluate these settings both with single task fine-tuning, and with multi-task training.
The results are given in Table \ref{table:image-rep}.

Faster R-CNN features are superior on VQA, which requires fine-grained localization
of objects in order to answer questions, while \resnext~ features are superior on Flickr30k and
Image Chat, which require a wide array of commonsense knowledge of different scenes.
On average across the tasks (last column), however, they provide similar performance.
As they provide different qualities, they are a good candidate for combination.
We thus provide both as input to our model and obtain superior single-task results on COCO, Flickr30k and VQA, with results on Image Chat as good as with \resnext~ and better than with Faster R-CNN. 
Multi-tasking performance also improves over previous results.
We thus adopt this combination strategy in subsequent experiments.

\paragraph{Multimodal Combiner}

We next assess the impact of the multimodal combiner module in our architecture; we first analyze the non-attentive version.
We consider either using it, or replacing it with a simple sum over feature type representations, see Section \ref{sec:mmc}.
We compare these alternatives on three tasks: COCO, Flickr30k and Image Chat, and examine performance both for our best performing combination features as well as for \resnext~alone.
The results are given in Table \ref{table:mix-rep}.
We see that without this component of the architecture,
the model can still give somewhat reasonable performance. 
However, by combining modalities with a Transformer architecture
we do see
improvements across all tasks. The MMC module takes as input
a sequence-based representation of the context history (token-level representation). We also experimented with 
giving the mean sequence representation  as input to the MMC instead, which
gave worse results (5\% regression on IC).
We thus report subsequent experiments using the full combiner using sequence-based inputs.

\paragraph{Freezing versus Fine-Tuning Encoders}

We compare the performance of our models when either freezing the image and text encoders after pre-training, or fine-tuning them in addition to the multimodal combiner of the language and vision tasks. If they are frozen, only the multimodal combiner is trained. Table \ref{table:freeze} presents the results,
comparing multi-task performance across several of our tasks. There are
clear wins from fine-tuning the encoders on the multimodal training data.

\begin{table}[h!]
\begin{small}
\begin{center}
\begin{tabular}{c|cccccc}
\hline
        & COCO           & Fl30k         & PC             & IC            & ICQA    & VQA            \\ \toprule
Freeze   & 27.9 & 57.2 & 40.6          & 40.6         & 37.6          & 64.5         \\ 
Fine-tune & 51.2          & 81.7 & 58.0 &  55.2      & 49.9     & 66.4  \\
\bottomrule
\end{tabular}
\end{center}
\end{small}
\caption{Training TransResNet-MMC on all tasks with freezing or not the text and image encoders.
\label{table:freeze}
}
\end{table}

\paragraph{Ranking vs. Classification Head}

We compare the performance of training VQA with the classification and
ranking heads, or training both at the same time. The results are shown in
Table \ref{table:heads}.

Training with a classification head alone (first row)
provides the best performance on VQA.
However, transfer to other tasks, shown by evaluating them using the ranking head, gives poor results, understandably as that has not been trained for (see Table \ref{table:main-res}, row 4).
Using a ranking head to train VQA gives far worse performance on VQA. 
We attribute this to
the subsampling of negative candidates in the loss function, rather than considering $\sim$3k possible candidates at once in the classification head.
The last row of Table \ref{table:heads} shows the performance of training both heads at once, and 
then evaluating the two heads. This dramatically improves the performance of 
the ranking head on VQA, as the classification head helps the model attain good weights. 
 
\begin{table*}[t!]
\begin{center}
\begin{small}
\begin{tabular}{cl|c|c|c|c|c|c|l|l}
Model & {\em Training data} &   COCO  & Flickr30k  & PC & IC & ICQA  & IGCQ & IGCQA & VQA  \\ \cline{1-10}
\multirow{12}{*}{\setstackgap{L}{1.5ex}\shortstack[l]{Existing \\ Models}}
& SCAN \cite{lee2018scan} & 50.4\textsuperscript{*} & 67.4 & - & - &- &- &- & -  \\
& SCG \cite{shi2019scg} & 56.6\textsuperscript{*} & 71.8 & - & - &- &- &- & -   \\
& Unicoder-VL \cite{li2019unicoder}& 62.3\textsuperscript{*} & 86.2 & - & - &- &- &- & -\\
& Unicoder-VL w/o pre-training & -  & 73.0 & - & - &- &- &- & - \\
& UNITER Base & 63.3\textsuperscript{*} & 84.7 & - &- &- &- & - & 72.3\textsuperscript{*} \\
& UNITER Large & 66.6\textsuperscript{*} & 88.2 & - &- &- &- & - & 73.2\textsuperscript{*} \\
& HDC \cite{nguyen2019multi} & 42.2 & 71.6 & - &- &- &- & - & 69.3\textsuperscript{*} \\
& VisualBERT (ST) \cite{li2019visualbert}   & -&-&-&-&-&-&-& 70.8 \textsuperscript{*} \\
& VisualBERT (ST) w/o pre-training   & -&-&-&-&-&-&-& 70.2 \textsuperscript{*} \\
& ViLBERT (ST) \cite{lu2019vilbert}   & -&-&-&-&-&-&-&70.6\textsuperscript{*} \\
& ViLBERT (ST) w/o pre-training  &-&-&-&-&-&-&-& 69.0\textsuperscript{*}  \\
& Pythia \cite{pythia01}\footref{pythiamodellink}  & -&-&-&-&-&-&-& 66.7 \\
& Pythia \footref{pythiamodellink} & -&-&-&-&-&-&-& 69.2\textsuperscript{*} \\
& ViLBERT (MT) \cite{lu201912in1} &-&-&-&-&-&-&-& 72.6\textsuperscript{*}\\
& ViLBERT (MT + FT) &-&-&-&-&-&-&-& 73.2\textsuperscript{*} \\
& TransResNet \cite{shuster2018engaging} & 44.3\textsuperscript{*} & 68.4  & 53.5 & 50.3 & 49.2 & 21.7 & 22.4 & - \\
\bottomrule
\end{tabular}
\end{small}
\end{center}
\caption{Previous state-of-the-art results. \textsuperscript{*} indicates results achieved by training with some or all of the validation set added to the train set, whereas we only train on the train set.  
Note that the results in this table correspond to a {\em different} model for each column, as the architectures are fine-tuned on each task separately rather than training a single architecture in a multi-task way. The ViLBERT (MT) model is a multi-task model, but uses image retrieval settings
on COCO and Flickr30k that are not comparable to the results presented here. The Pythia models on rows 9 and 10 are the same except they are trained with the VQA train set and VQA train + valid set respectively, thus we list both numbers. 
\label{table:previous-work}
}
\end{table*}

\begin{table*}[t!]
\begin{center}
\begin{small}
\begin{tabular}{cl|c|c|c|c|c|c|c|c|c}
 Arch. & {\em Training data} &   COCO  & Flickr30k  & PC & IC & ICQA  & IGCQ & IGCQA & VQA & Avg \\ \cline{1-11}
MMC & ST COCO & 57.2 & 69.4  & 24.0 & 16.5 & 13.1 & 13.4 & 10.3 & 0.3 & 25.5 \\ 
MMC & ST Flickr30k & 27.7 & 79.7 & 23.0 & 16.3 & 13.8 & 15.8 & 12.2 & 0.2 & 23.6 \\ 
MMC & ST IC & 20.0  & 40.5 & 57.3 & 56.3 & {\bf 55.2} & \textbf{35.8} & \textbf{43.3} & 0.4 & 38.6\\ 
MMC & ST VQA    & 0.0  & 0.3 & 1.2  & 1.3  & 1.2 & 1.6 & 1.9 & \textbf{67.0} & 9.3  \\ 
MMC & MT + FT & \textbf{59.6} & \textbf{84.0} & 58.8 & 55.9 & 51.4 & 30.2 & 41.1 & 66.5 &  \color{gray}{\em 56.0}\\
\cline{1-11}
MMC & MT & 51.2 & 81.7 & 58.0 & 55.2 & 49.9  & 25.7 & 38.4 & 66.4 & 53.3\\
2AMMC & MT  & 54.2 & 82.0 & {\bf 59.5} &  \textbf{56.9} & 52.3 & 28.1 & 38.1 & 65.6 & 54.6\\
3AMMC & MT  & 52.7 & 82.9 & 58.5 & 56.1 & 52.4  & 31.4 & 39.8 &66.9 & \textbf{55.1} \\
4AMMC & MT  & 53.2 & 81.8 & 58.7 & 56.2 & 54.5 & 31.8 & 35.8 &  65.9 & 54.8 \\
\bottomrule
\end{tabular}
\end{small}
\end{center}
\caption{ Multi-tasking test results of our models. The first four rows show the transfer performance of our TransResNet-MMC model trained on a single task (ST), indicated in the Training data column.
The fifth row shows a multi-task  model which is then fine-tuned (MT+FT) on 
each single task separately (each column corresponds to a separate model, 
we hence report average performance in gray italics).
The bottom four rows compare performance of single multi-task models with different types of
multimodal combiners. The multi-task performance is close to single-task performance, and in some cases better across several tasks. The attentive multimodal combiner (AMMC)  obtains the best overall average performance.
\label{table:main-res}
}
\end{table*}

\paragraph{Single Task Results}

Using our best approach, we now report final results fine-tuned on each task independently.
The results are given in Table \ref{table:main-res}.
We report results across all evaluation sets for a given training target.
E.g., the first row shows the model performance when training with COCO, evaluated on the test sets of COCO, Flickr30k, Personality Captions, Image Chat, Image Chat QA, VQA, IGCQ, IGCQA and VQA.
As expected, we observe better results on the test set of the task being trained on than on other test sets. However, there is some transfer between some of the tasks.
For example, training on COCO gives non-trivial performance on Flickr30k, and vice-versa, although Flickr30k training helps less, probably due to its smaller size.

\paragraph{Multi-Task Results}

Results of our Multi-Task models are given in Table \ref{table:main-res}, last four rows. We first assess the performance of the MMC MT model (without an attentive multimodal combiner).
We achieve marginally superior performance on Personality Captions and Flickr30k,
and marginally inferior performance on the other tasks compared to our best single-task (ST)
models, but in a single conversational agent. The final column showing average performance across tasks
makes this point clear, as those numbers are vastly superior for the multi-task models.
Like our single-task counterparts,
many of these results are still well above previous state of the art, e.g. on  Personality Captions and Image Chat, and within range of the state of the art on COCO and Flickr30k.

\paragraph{Multi-Task Results with Attentive Multimodal Combiner}
We next assess the effect of Multi-Task training with multiple Transformers in the multimodal combiner (2, 3 or 4  instead of the single Transformer in our base architecture).  The bottom four rows in Table~\ref{table:main-res} show that 
using the attentive multimodal combiner leads to improvements in average performance over all tasks, with 2AMMC achieving the best results on PC and IC tasks of all methods, and 3AMMC being slightly better on average.
Note that the early stopping criterion for
these experiments is the average
performance over all tasks, which
leads to performance gains shifting
between tasks among architectures, while the average itself is controlled.
This could be altered by selecting a different stopping criterion, as detailed further below and in Table~\ref{table:mt-st-earlystop}.
Table~\ref{table:ammc-breakdown-res} breaks down the performance obtained on all tasks by each of the Transformers in the 3AMMC.
There are striking differences between the tasks as to how performance is split among the three MMCs: on VQA, MMC-1 and MMC-2 have near 0 performance while MMC-3 performs as well as the full system, but this comes at the expense of much worse performance on all the conversational tasks compared to MMC-1 and MMC-2.
On PC, MMC-1 performs nearly as well as the full system and much better than MMC-2 and MMC-3. The overall best performance on all other tasks requires combining all three MMCs.
To see if the performance gains of AMMC come just from the network being larger, we compare to MMC modules with more layers up to an equivalent size, see Table \ref{table:ammc-size-compare}. The results show that making standard MMC larger only hurts performance. Increasing the number of MMC heads similarly degrades performance (results not shown). These results highlight the benefits of the AMMC design.

\begin{table*}[t!]
\begin{center}
\begin{small}
\begin{tabular}{cl|c|c|c|c|c|c|c|c|c}
 Arch. & &   COCO  & Flickr30k  & PC & IC & ICQA  & IGCQ & IGCQA & VQA & Avg \\ \cline{1-11}

3AMMC (MMC-1) & MT & 24.7 & 61.6 & 48.9 & 45.1 & 27.8  & 24.0 & 22.1 & 1.1 & 32.0\\
3AMMC (MMC-2) & MT  & 31.1 & 50.6 & 19.5 & 26.0 & 27.8 & 27.4 & 33.5 &0.0 & 27.0 \\
3AMMC (MMC-3) & MT & 31.0 & 61.9 & 21.3 & 13.0 & 9.5 & 8.9 & 13.0 &   66.9 & 28.2 \\
\bottomrule
\end{tabular}
\end{small}
\end{center}
\caption{Results on each dataset when we evaluate our 3AMMC model by only taking a single MMC output as the context representation. The first MMC alone already gives good performance on PC and IC, and the third on VQA. All three are needed for some of the tasks.
\label{table:ammc-breakdown-res}
}
\end{table*}

\begin{table*}[t!]
\begin{center}
\begin{small}
\begin{tabular}{ccl|c|c|c|c|c|c|c|c|c}
 MMC Arch. & (Compare to) & &   COCO  & Flickr30k  & PC & IC & ICQA  & IGCQ & IGCQA & VQA & Avg \\ \cline{1-11}

4 Layers &(2 AMMC) & MT & 51.4 & 81.0 & 56.0 & 53.3 & 48.5  & 29.1 & 39.9 & 66.7 & 53.3\\
6 Layers &(3 AMMC) & MT  & 48.6 & 78.1 & 57.1 & 53.4 & 49.7 & 30.8 & 36.8 & 66.1 & 52.6 \\
8 Layers &(4 AMMC) & MT & 35.6 & 65.6 & 36.2 & 33.6 & 31.8 & 26.6 & 26.8 &   59.0 & 39.4 \\
\bottomrule
\end{tabular}
\end{small}
\end{center}
\caption{Test results when we train our MMC models with varying numbers of layers, which we compare to our AMMC model sizes. Increasing the number of MMC layers only hurts performance. 
\label{table:ammc-size-compare}
}
\end{table*}

\begin{table*}[t!]
\centering
\begin{adjustbox}{center=15.5cm}\setlength{\tabcolsep}{0.2em}
\begin{small}
\begin{tabular*}{\textwidth}{ccl}
\hline
\small{Image} &  & \small{Output}  \\  \hline
\hline
\\[-1.8ex]
\multirow{4}{*}{\includegraphics[height=11ex, width=19ex]{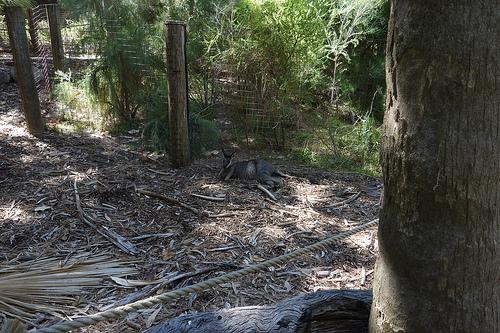}}
\\[-1.5ex]
& \bf{\small{Task}} & \small{Coco}  \\
& & \\
& & \\
& \bf{\small{TransResNet MMC}} & \small{there is a broken tree log on the ground} \\[0.5ex]
\hline
\\[-1.8ex]
\multirow{4}{*}{\includegraphics[height=11ex, width=19ex]{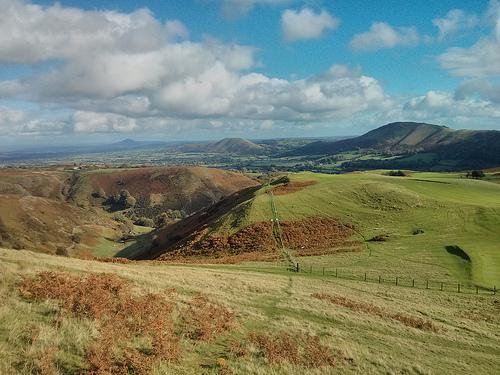}}
\\[-1.5ex]
& \bf{\small{Task}} & \small{Coco}  \\
& & \\
& & \\
& \bf{\small{TransResNet MMC}} & \small{A large grass covered field under a mountain.} \\[0.5ex]
\hline
\\[-1.8ex]
\multirow{4}{*}{\includegraphics[height=11ex, width=19ex]{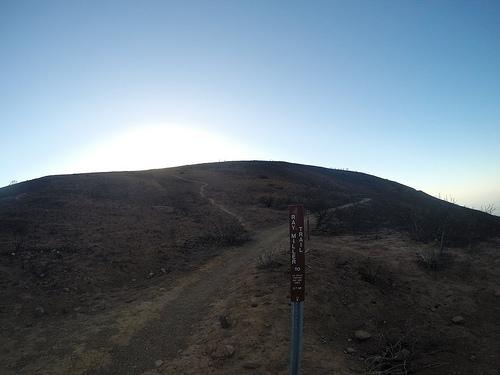}}
\\[-1.5ex]
& \bf{\small{Task}} & \small{Flickr30k}  \\
& & \\
& & \\
& \bf{\small{TransResNet MMC}} & \small{ A chaparral landscape scene void of human residence.} \\[0.5ex]
\hline
\\[-1.8ex]
\multirow{4}{*}{\includegraphics[height=11ex, width=19ex]{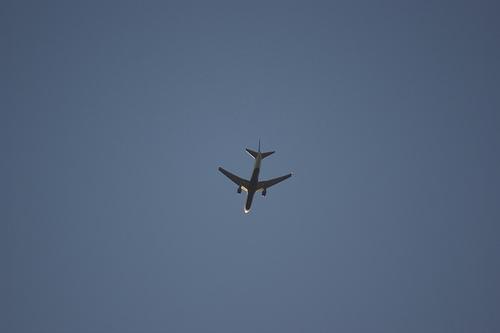}}
\\[-1.5ex]
& \bf{\small{Task}} & \small{Flickr30k}  \\
& & \\
& & \\
& \bf{\small{TransResNet MMC}} & \small{A plane flying sideways.} \\[0.5ex]
\hline
\\[-1.8ex]
\multirow{4}{*}{\includegraphics[height=11ex, width=19ex]{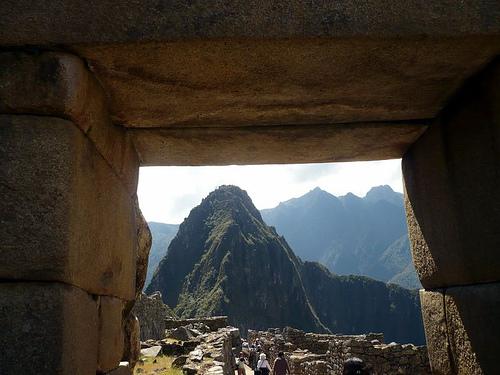}}
\\[-1.5ex]
& \bf{\small{Task}} & \small{VQA}  \\
& \bf{\small{Context}} & \small{What is the color of the mountain?}  \\
& & \\
& \bf{\small{TransResNet MMC}} & \small{gray} \\[0.5ex]
\hline
\\[-1.8ex]
\multirow{4}{*}{\includegraphics[height=11ex, width=19ex]{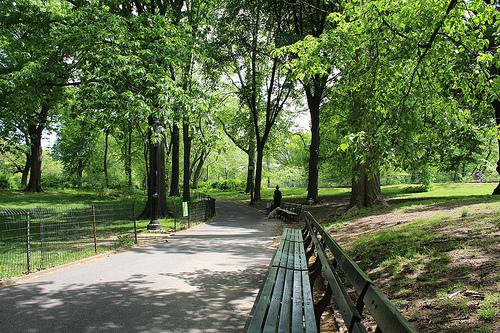}}
\\[-1.5ex]
& \bf{\small{Task}} & \small{VQA}  \\
& \bf{\small{Context}} & \small{Does it appear to be rainy?}  \\
& & \\
& \bf{\small{TransResNet MMC}} & \small{no} \\[0.5ex]
\hline
\\[-1.8ex]
\multirow{4}{*}{\includegraphics[height=11ex, width=19ex]{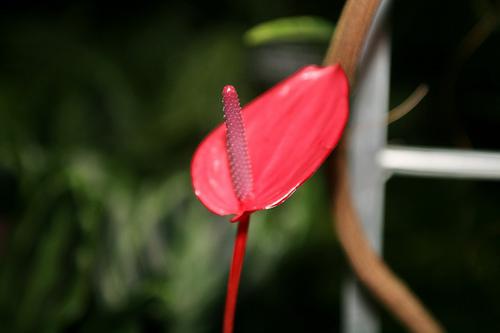}}
\\[-1.5ex]
& \bf{\small{Task}} & \small{Personality Captions (Style: Happy)}  \\
& & \\
& & \\
& \bf{\small{TransResNet MMC}} & \small{Wow what a beautiful and perfect shade of pink and red! I am so captivated!} \\[0.5ex]
\hline
\\[-1.8ex]
\multirow{4}{*}{\includegraphics[height=11ex, width=19ex]{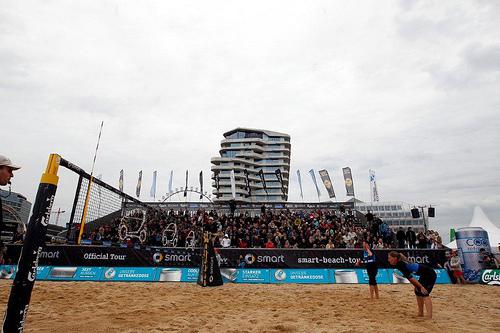}}
\\[-1.5ex]
& \bf{\small{Task}} & \small{Personality Captions (Style: Attractive)
}  \\
& & \\
& & \\
& \bf{\small{TransResNet MMC}} & \small{Wow I would love for someone to create this setting in the sand for me.
} \\[0.5ex]
\hline
\\[-1.8ex]
\multirow{4}{*}{\includegraphics[height=11ex, width=19ex]{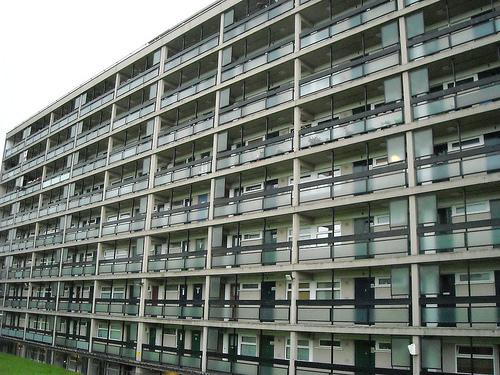}}
\\[-1.5ex]
& \bf{\small{Task}} & \small{Image Chat (Style: Compassionate)}  \\
& & \\
& \bf{\small{Context}} & \small{Round 1: Something about the pattern calms me.}  \\
& \bf{\small{TransResNet MMC}} & \small{The architecture calms you.} \\[0.5ex]
\hline
\\[-1.8ex]
\multirow{4}{*}{\includegraphics[height=11ex, width=19ex]{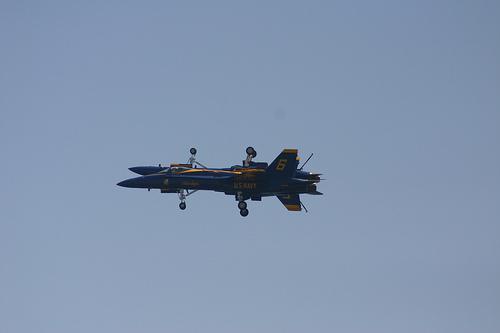}}
\\[-1.5ex]
& \bf{\small{Task}} & \small{Image Chat (Style: Emotional)}  \\
& \bf{\small{Context}} & \small{Round 1: Airplanes are scary to get on, you never know if it will crash or not.} \\ & & \small{Round 2: But these are professional pilots though. }  \\
& \bf{\small{TransResNet MMC}} & \small{They are, and for many people they mean a lot. My grandfather loved planes!} \\[0.5ex]
\hline
\hline

\hline
\\[-1.8ex]

\hline
\end{tabular*}
\end{small}
\end{adjustbox}
\caption{Example output from our TransResNet MMC multi-task model for different tasks.
\label{fig:example_preds}
}
\end{table*}

\paragraph{Multi-Tasking Small vs. Large Tasks}

The tasks we do see a performance gain in when multi-tasking, Flickr30k and Personality-Captions, tend to be smaller tasks where there is a larger related task also in the multi-tasking set, in this case COCO and Image Chat.
To investigate the effects of training set size on multi-tasking transfer we thus conducted experiments to see if we observe the same effects of improvement on another dataset if we downsampled it. We thus consider adjusting the training set 
size of COCO to be the same size as Flickr30k, and then consider multiples of that size, and observe the change in performance with changing size. We  compare single-task
training on that subset to multi-task training with all other tasks and that subset.
For these experiments we considered a smaller hyperparameter sweep for simplicity, with a multimodal combiner of 2 layers and sweep across different number of heads for the multi-head attention, explaining the slightly lower results.
We perform early stopping on COCO. 
The results are given in Table \ref{table:mt-coco-sizes}. 
We observe for single-task training a drop from 54\% accuracy to 42.1\% as we go from 83k examples down to 29k. Multi-tasking with the full COCO dataset also yields the same 54\% accuracy, which makes it appear that multi-tasking is not useful for generalization. 
However, subsampling COCO reveals a different story -- the smaller the training set, the more the multi-tasking helps, with a gap of 42.1\% to 49.3\% in the 29k training example case. As researchers who construct new tasks often collect large scale datasets, this means multi-tasking will often have less effect than is observed in a few-shot setup. 

\begin{table}[h!]
\begin{center}
\begin{tabular}{l|cc}
\hline
COCO Train Size         & Multi-Task    & Single-Task    \\
\toprule
1.0x Flickr30k  (29000)   & 49.3 & 42.1 \\
1.5x Flickr30k  (43500) & 51.6 & 50.3  \\ 
2.0x Flickr30k (58000)        & 53.7 & 51.9  \\ 
2.5x Flickr30k (72500)      & 53.8 & 53.6  \\ 
Full Size (82783)       & 54.0 & 54.0    \\ 
\bottomrule
\end{tabular}
\end{center}
\caption{Accuracy on COCO test set when downsampling COCO during training to the same size as the Flickr30k training set, or multiples thereof. Smaller training sets are clearly helped by multi-tasking. Eventually there is enough data of the single task.
\label{table:mt-coco-sizes}
}
\end{table}

\paragraph{Multi-Tasking + Single-Task Fine-Tuning}

While our goal is to make a single agent that is good at all our tasks, we also investigate if multi-tasking can help improve performance on a single task, by either multi-tasking and early stopping on a particular task,  or multi-tasking and then fine-tuning on a particular task. 

Early stopping test results are shown in Table \ref{table:mt-st-earlystop}.  We report for each task out of COCO, Flickr30k, Image Chat and VQA the performance on the test set of that task itself, as well as transfer performance to the other tasks. These results can be compared to the results of optimizing multi-task performance in the last row "Avg.", see also Table \ref{table:main-res} (sixth row).
 There are clear gains for each task that is early stopped, but at large expense for the other tasks. For example fine-tuning on COCO gives 54.0\% compared to 51.2\% when multi-tasking, but is still worse than the 57.2\% when training as a single-task.  
  Transfer to Flickr30k is still good, likely as they are similar tasks, but Image Chat results are then poor.
 On Flickr, the early stopping result of 83.0\% is superior to both the multi-task result
 of 81.7\% and the single-task result of 79.7\%. This can be explained by Flickr30k being smaller than COCO, and thus benefiting from multi-tasking more, as we explained in the previous section.

Multi-tasking followed by single task fine-tuning test results are shown in Table \ref{table:mt-plus-ft} (also summarized in Table \ref{table:main-res}). Generally, these are superior to the multi-tasking per-task early stopping results. For example 
fine-tuning on COCO gives 59.6\% compared to 54.0\% when multi-tasking and early stopping, or even 57.2\% when training as a single-task, so it is the best result we obtain over all methods.  We also achieve our best results in this fashion on Flickr30k. For VQA the validation results were higher (not shown) in this setting, but resulted in slightly inferior test numbers, so a small amount of overfitting occurs.

\paragraph{Comparison to Existing Results}
We give results from previous work in Table \ref{table:previous-work}.
Our results compare favorably on the conversational tasks, i.e. PC, IC, ICQA, IGCQ and IGCQA.
For the COCO, Flickr30k and VQA tasks, our results are within range of the state of the art, but are surpassed 
by some of the methods.
We note that on COCO others used the validation set for training 
whereas we did not (see Sec. \ref{sec:vltasks}, we do not want multi-task experiment train and valid data to overlap).
 For VQA we report the number from Pythia\footnote{\url{https://learnpythia.readthedocs.io/en/latest/tutorials/pretrained_models.html##pretrained-models}\label{pythiamodellink}} as a comparison point, 
 as that  method uses the train set only without VQA data augmentation from the Visual Genome, VisDial or other data augmentations (similar to us) and we used their setup as a starting point for our implementation.
 Our numerical results are comparable to theirs.

\paragraph{Larger-Scale Cross-Module Pre-training} 
Some of the best-performing methods on a subset of tasks rely on large-scale cross-module pre-training \cite{chen2019uniter,li2019unicoder,lu2019vilbert}, which leads to better performance but requires gigantic multimodal datasets like Conceptual Captions \cite{sharma2018conceptual} or Visual Genome \cite{krishnavisualgenome}, as shown in Table~\ref{table:compare-training-resources}, as well as high computing resources (e.g., 882 and 3645 V100 GPU hours for UNITER-base and UNITER-large, respectively).
Pre-training on COCO alone as done in \cite{li2019visualbert} gives more limited improvement (see Table~\ref{table:previous-work}).
Our approach combines 
vision and text encoders with minimal additional multimodal training resources required. Even counting all the multi-task datasets used for training adds up to only 1M image-sentence (I-S) pairs, resulting in training that takes around 40 V100 GPU hours.
We expect larger-scale cross-module pre-training would also improve the performance of our models, but this is beyond the scope of this work.

\begin{table}[h!]
\begin{center}
\begin{tabular}{l|cc}
\hline
Model & Dataset & Size (I-S Pair)   \\
\toprule
UNITER   & COCO,VG,CC,SBUC&  9.6 M \\
ViLBERT  & CC & 3.0 M \\
Unicoder-VL & CC, SBUC &  3.8 M \\ 
\bottomrule
\end{tabular}
\end{center}
\caption{Sizes of multimodal pre-training datasets in terms of image-sentence pairs. Our model obtains comparable results on all tasks without any cross-module pre-training on large datasets such as Visual Genome (VG), Conceptual Captions (CC), or SBU Captions (SBUC). Thus, multi-tasking can be viewed as a strong alternative to large-scale pre-trainining, considering its simplicity and effectiveness in terms of computation power. 
\label{table:compare-training-resources}
}
\end{table}

\paragraph{Example Predictions}

We show example predictions of our MMC multi-task model in Table \ref{fig:example_preds}. We take test images, and for COCO, Flickr30k, Personality Captions and Image Chat we show the top ranked candidate using the ranking head, ranking all utterances from the given training set.
For VQA we show the output of the classification head.
We observe that the same underlying model can produce a diverse
range of  outputs depending on the task, ranging from factual captioning to
conversations grounded on the image.

\section{Conclusion}

In order to build an image-grounded conversational agent, 
we have assembled disparate multimodal tasks and built a single
architecture for multi-task training on them, incorporating 
a novel attentive multimodal combination module.
Through detailed analysis and ablations, we have shown that 
our approach can obtain strong performance across a number of tasks.
Future work could 
investigate further how these skills are blended during interaction,
rather than evaluate them as stand-alone tasks, and consider more tasks.

\section{Acknowledgements}
We are grateful to Amanpreet Singh and Vedanuj Goswami for providing help and advice, comparison results and faster R-CNN image features. We also thank Mary Williamson and Eric Smith for very useful discussions.

\FloatBarrier

\bibliographystyle{named}
\bibliography{ijcai20}

\end{document}